\documentclass[5p,times]{elsarticle}

\makeatletter
\def\ps@pprintTitle{%
  \let\@oddhead\@empty
  \let\@evenhead\@empty
  \let\@oddfoot\@empty
  \let\@evenfoot\@oddfoot
}
\makeatother

\usepackage{amsmath}
\usepackage{subcaption}
\usepackage{booktabs}
\usepackage{dblfloatfix}
\usepackage{multirow}
\usepackage{enumitem}

\usepackage{array}
\newcolumntype{x}[1]{>{\centering\arraybackslash\hspace{0pt}}p{#1}}

\journal{Future Generation Computer Systems}

\begin{document} \sloppy

\begin{frontmatter}

\title{PROMPT: Learning Dynamic Resource Allocation Policies\\ for Network Applications}

\author[label1]{Drew Penney}
\author[label2]{Bin Li}
\author[label3]{Jaroslaw J. Sydir}
\author[label1]{Lizhong Chen\corref{cor1}}
\author[label2]{Charlie Tai}
\author[label1]{Stefan Lee}
\author[label4]{Eoin Walsh}
\author[label4]{Thomas Long}

\address[label1]{School of Electrical Engineering and Computer Science, Oregon State University, Corvallis, OR, USA}
\address[label2]{Intel Labs, Hillsboro, OR, USA}
\address[label3]{Intel Labs, Santa Clara, CA, USA}
\address[label4]{Intel Corporation, Shannon, Ireland}

\cortext[cor1]{Corresponding author. Email address: chenliz@oregonstate.edu}

\begin{abstract}
A growing number of service providers are exploring methods to improve server utilization and reduce power consumption by co-scheduling high-priority latency-critical workloads with best-effort workloads. This practice requires strict resource allocation between workloads to reduce contention and maintain Quality-of-Service (QoS) guarantees. Prior work demonstrated promising opportunities to dynamically allocate resources based on workload demand, but may fail to meet QoS objectives in more stringent operating environments due to the presence of resource allocation cliffs, transient fluctuations in workload performance, and rapidly changing resource demand. We therefore propose PROMPT, a novel resource allocation framework using proactive QoS prediction to guide a reinforcement learning controller. PROMPT enables more precise resource optimization, more consistent handling of transient behaviors, and more robust generalization when co-scheduling new best-effort workloads not encountered during policy training. Evaluation shows that the proposed method incurs 4.2x fewer QoS violations, reduces severity of QoS violations by 12.7x, improves best-effort workload performance, and improves overall power efficiency over prior work.
\end{abstract}






\end{frontmatter}


\section{Introduction}\label{sec:introduction}

User-facing applications, such as e-commerce and videoconferencing, represent a critical and growing class of workloads for service providers. Performance of these high-priority (HP) workloads is often synonymous with user experience, so service providers usually specify strict Quality-of-Service (QoS) requirements. Conventional approaches to meet these requirements, such as executing these HP workloads in isolation (i.e., one workload per machine), can ensure satisfactory performance, but are undesirable due to substantial over-provisioning during periods of low demand. More recently, service providers have sought to ameliorate these adverse effects by opportunistically co-scheduling best-effort (BE) workloads, thereby enabling higher average resource utilization. Co-scheduling can, however, introduce substantial resource contention between HP and BE workloads that may compromise user experience.

Prior work has explored a variety of methods to limit resource contention by strictly managing the resource allocation for each workload. Simple approaches, such as statically allocating resources to the HP workload based on peak demand, are usually highly inefficient since average demand is often substantially lower than peak demand \cite{above_the_clouds_2009, chen2014consolidating}. Incremental improvements are possible by adjusting resources on a time-of-day basis, although deviations from historic averages may compromise QoS guarantees \cite{gong2010press, amiri2017survey}. For these reasons, recent work has explored dynamic resource allocation as a more adaptable solution. These approaches vary; some exploit workload-specific knowledge \cite{Ubik2014, Dirigent2016}, several apply general search-based methods \cite{Heracles2015, Parties2018, Adaptive_QoS_Pred_2019, CLITE2020, MOBO-NFV} or tabular reinforcement learning \cite{Hipster2017, RL_Prediction_2020}, and a few recent works consider deep reinforcement learning \cite{RLDRM2020, Twig2020}. In general, these dynamic methods provide promising improvements in BE performance, but may compromise QoS in more stringent operating environments as they cannot appropriately accommodate QoS cliffs, transient fluctuations in workload performance, and rapidly changing resource demand. Addressing these limitations requires innovative solutions that can proactively and precisely predict the effects of changing and even unseen workload demands.

In this paper, we propose \textbf{P}redictive \textbf{R}esource \textbf{O}ptimization for \textbf{M}ultiple \textbf{P}rioritized \textbf{T}asks (PROMPT), a novel machine learning framework for dynamic resource allocation. PROMPT is designed around proactive QoS predictions in order to eliminate the fundamental reliance on QoS measurements observed in prior work. These QoS predictions, derived from general resource contention indicators, are shown to provide more fine-grained insight into potential worst-case behaviors and more consistent feedback for a reinforcement-learning-based resource allocation controller. Furthermore, while many prior works assume a predominantly fixed set of workloads \cite{RLDRM2020, Twig2020}, PROMPT is designed to handle more general operating conditions in which co-scheduled BE workloads may change frequently, thereby avoiding any sacrifices in HP QoS that could be introduced by online search or policy re-training. Evaluations on a real-world networking platform show that PROMPT incurs 4.2x fewer QoS violations, reduces the severity of QoS violations by 12.7x, improves BE performance, and improves overall power efficiency compared with prior work.

The rest of the paper is organized as follows: Section 2 provides background on reinforcement learning and related work on dynamic resource allocation; Section 3 describes several key challenges in dynamic resource allocation that are not addressed by prior work; Section 4 details our proposed resource allocation framework; Section 5 describes methodology and evaluation; Section 6 concludes.

\begin{figure}[t]
\centering
\includegraphics[width=0.37\textwidth]{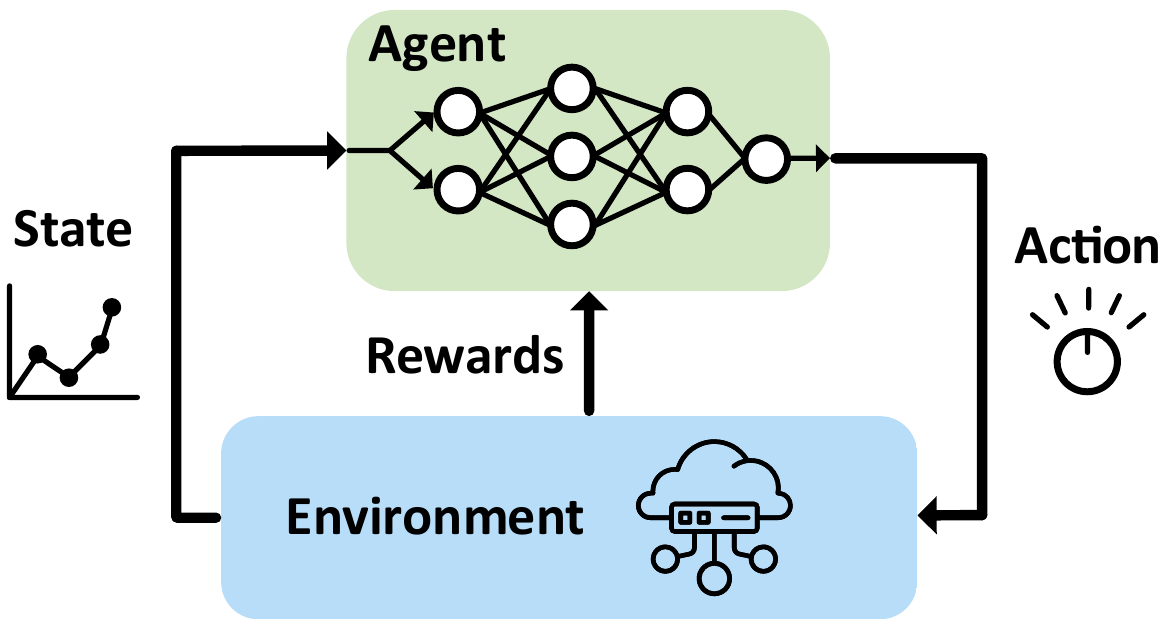}
\vspace{-0.15cm}
\caption{Generic reinforcement learning framework.}
\vspace{-0.15cm}
\label{fig:RL}
\end{figure}

\begin{figure*}[t]
    \centering
    \begin{subfigure}[b]{0.4\linewidth}%
        \centering
        \includegraphics[width=0.9\linewidth]{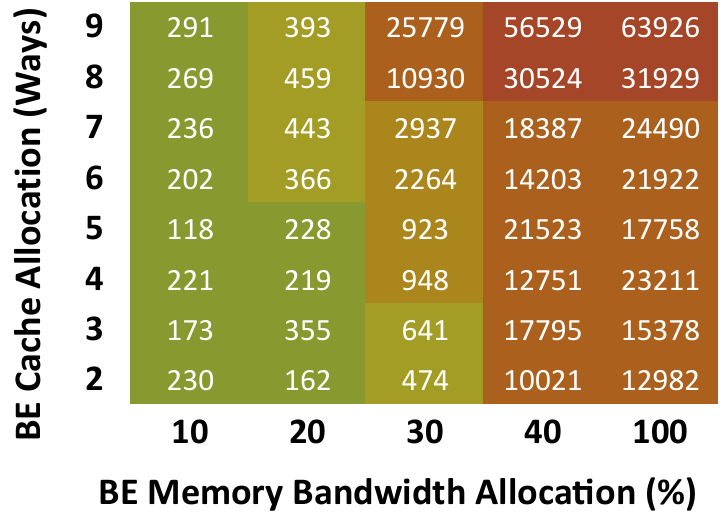}
        \caption{Maximum observed HP packet drop rate (i.e., dropped packets per second) across all one-second intervals in a 100-second period.}
    \end{subfigure}%
    \hfill
    \begin{subfigure}[b]{0.55\linewidth}%
        \centering
        \includegraphics[width=0.9\linewidth]{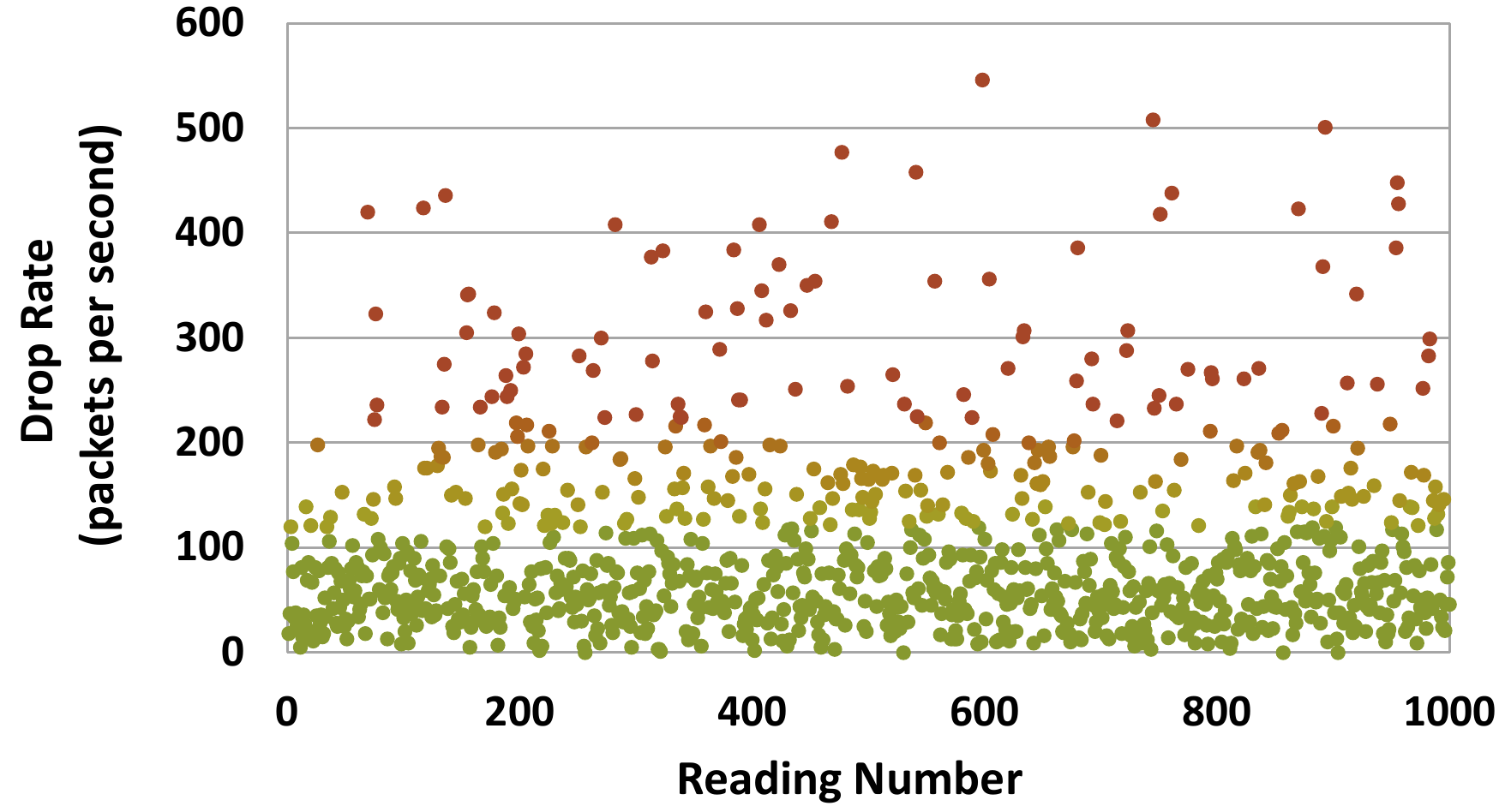}
        \caption{HP QoS measurements at one second sampling interval.}
    \end{subfigure}
    \caption{HP QoS behavior when co-scheduling a HP networking workload with a BE machine learning workload. HP workload demand is held at a fixed rate using an external traffic generator.}
    \label{fig:coscheduling_behaviors}
\end{figure*}

\section{Background \& Related Work} \label{sec:related_work}

\subsection{Reinforcement Learning}
 Reinforcement learning is a branch of machine learning founded upon self-guided learning and optimization of problems that can be modeled as a sequence of control decisions \cite{sutton2018reinforcement}. Fundamental to this learning process are the specifications for state, action, and reward, as illustrated in Figure \ref{fig:RL}. The environment defines the task that we seek to optimize, which corresponds to the compute node and its resource allocation. The \textit{state} $s_t$ represents the observable information about our environment, such as workload behaviors and current resource utilization. This state information is passed to the agent, representing the decision-maker, which continuously interacts with the environment by taking actions $a_t$. Finally, rewards $r_t$ are given as a signal to either encourage or discourage changes in the environment resulting from prior actions. In the context of this paper, the overall goal of the agent is to learn a resource allocation policy $\pi^*$ that maximizes cumulative rewards, with discounting over time ($\gamma$), serving as a proxy for user-defined goals such as workload QoS (Equation \ref{eq:OptPolicy}).

\begin{equation}
\label{eq:OptPolicy}
  \pi^* = arg \max_\pi\mathbb{E} [\sum_{t\geq0}\gamma^t*r_t |  s_0=s, \pi].
\end{equation} 

\subsection{Related Work}

Applications for resource allocation optimization span diverse domains, ranging from wireless communication networks to smart electric grids. Naturally, each domain introduces its own unique challenges in applying resource allocation optimization, thus the need for continued research.

One broad category of work focuses on cluster-level schedulers that assign workloads to compute nodes \cite{quincy2009, delayschedule2010, qclouds2010, bubbleup2011, paragon2013, omega2013, wharemap2013, quasar2014, towards2014, smite2014, multiobjectiveplace2015, tarcil2015, borg2015, qaware2015, hcloud2016, deeprm2016, improving_spark_2017, hierarchical2017, qosfog2021, proactive_alloc_2021, cosco, gosh} and adapt these assignments based on either historical or predicted resource demand \cite{amiri2017survey, amiri2017ids, amiri2018online, amiri2018sequential, amiri2020new}. These cluster-level schedulers usually assume minimal and/or constant resource contention between co-scheduled workloads or assume that this resource contention will be handled by per-node, dynamic resource allocation controllers, which are the target of our work. 

Approaches for dynamic resource allocation in cloud applications have varied dramatically. Most early works were limited by available resource partitioning mechanisms, so were designed around relatively coarse-grained heuristics to reduce contention by temporarily pausing BE workload execution \cite{bubbleflux2013, cpi2_2013, maximizing2013, rubik2015}. Other works apply workload-specific knowledge \cite{Ubik2014, Dirigent2016}, which is generally not practical in cloud applications. More recent work can be divided into two general categories: search-based methods and reinforcement-learning-based methods. 

Search-based methods can be further decomposed into works using hill-climbing \cite{Heracles2015, Parties2018}, genetic algorithms \cite{Adaptive_QoS_Pred_2019}, and Bayesian optimization \cite{CLITE2020, MOBO-NFV}. Most search-based methods rely upon online search to determine appropriate resource allocations. This search can, however, lead to significant QoS violations in highly-dynamic operating environments (Section \ref{sec:results}). Concurrent research on HP networking workloads instead uses offline search to generate a lookup table for online, load-driven optimization, but does not support co-scheduling with BE workloads \cite{dpm-nfv}. Among reinforcement-learning-based methods, two are based on tabular Q-learning \cite{Hipster2017, RL_Prediction_2020} and two are based on deep Q-learning \cite{RLDRM2020, Twig2020}. Tabular methods cannot effectively scale to multiple workloads/resources and exhibit oscillatory control behaviors in practice \cite{Twig2020}. Different from our work, RLDRM \cite{RLDRM2020} assumed a static set of workloads, focused on allocation of a single resource (last-level cache), and used current workload demand and resource allocation as the state for the reinforcement-learning controller which, as discussed in Section \ref{sec:preliminaries}, greatly restricts handling of transient QoS violations and generalization to unseen BE workloads. Twig \cite{Twig2020} was instead designed to co-schedule HP workloads only so cannot be used with BE workloads and, similar to all prior work, is fundamentally reliant upon QoS measurements so cannot be used to mitigate transient QoS violations.

Another largely orthogonal category of work explores dynamic resource allocation for microservices (i.e., loosely coupled, intercommunicating HP workloads) \cite{seer2019, firm2020, rambo2021, sinan2021}. 
Those works primarily focus on specialized mechanisms to identify critical execution paths between microservices. In contrast, networking workloads described in this paper have very few execution paths per traffic direction (i.e., to/from the user), so would not meaningfully benefit from critical-path detection. Moreover, all paths in each direction may be executed as a single thread, which prevents path-level resource allocation. Works on microservices also generally do not consider resource contention, so would require a node-level resource allocation controller in order to co-schedule BE workloads. Lastly, we note that the QoS prediction mechanism introduced in Seer \cite{seer2019} is fundamentally limited due to its use of queue depth as the only information source, which exhibits the same problematic behaviors as direct QoS measurements (i.e., often zero and highly transient).

\section{Preliminaries}\label{sec:preliminaries}

\subsection{Problem Formulation}
In this paper, we consider an operating environment with two priority levels for workloads: 1) HP workloads with strict performance (i.e., QoS) targets and 2) BE workloads without strict performance targets. Formally, we denote these workloads $W=\{W_1^{HP},...,W_x^{HP}, W_1^{BE},...,W_y^{BE}\}$. Similarly, we denote HP workload QoS targets as $Q=\{Q_1^{HP},...,Q_x^{HP}\}$. Measured performance for workload $w$, which can vary over time $t$, is then denoted as $P_{meas}(W_{w,t})$. Each physical compute node on which workloads are co-scheduled generally has multiple resources, $R=\{R_1,...,R_n\}$, each with a finite number of discrete units to be allocated, denoted as $R^*=\{R_1^*,..., R_n^*\}$. At each timestep $t$, each workload $w$ must be allocated a portion of each resource $r$, denoted as $R_{r,w,t}$. The goal of workload co-scheduling, as given in Equation \ref{eq:coscheduling}, is to allocate these finite system resources in a manner that maximizes performance for BE workloads while satisfying QoS targets for all HP workloads. Note that the resource allocation required to satisfy HP QoS targets may continuously vary over time along with changes in HP workload demand (e.g., packets per second), thus the need for dynamic resource allocation.
\begin{equation}
    \begin{aligned}
    \text{maximize}   \qquad & \sum_{w=1}^y P_{meas}(W_{w,t}^{BE})                        \\
    \text{subject to} \qquad & P_{meas}(W_{w,t}^{HP}) \geq Q_w^{HP}   &&   \forall w=1,x  \ \ \& \ \ \forall t=1,T \\
                  \qquad & R_m^* \geq \sum_{w\in W} R_{r,w,t}   &&   \forall r=1,n \ \ \& \ \ \forall t=1,T  \\
    \end{aligned}
    \label{eq:coscheduling}
\end{equation}
%

\subsection{Challenges}

Workload co-scheduling using dynamic resource allocation offers distinct advantages over traditional static methods, but requires precise mitigation of resource contention to preserve QoS. This contention can manifest as diverse challenges, several of which have not been adequately addressed by prior work. We demonstrate these challenges by characterizing a prominent network-edge workload --- the virtual Broadband Network Gateway (vBNG). This workload serves an essential role as the access point for business, residential, and wholesale connectivity \cite{BNG_whitepaper}. Strict QoS goals must be maintained to ensure continuity of service. Note that the demonstrated behaviors are observed across a wide variety of workload co-scheduling scenarios, which we discuss further in Section \ref{sec:results}.


\subsubsection{QoS Cliffs}

Many workloads, even in isolation, exhibit dramatic performance degradation when resources are lowered beyond a critical threshold (i.e., QoS cliff). These effects become more pronounced when co-scheduling workloads due to contention over shared resources. As an example, Figure~\ref{fig:coscheduling_behaviors}(a) shows packet drop rate of an HP vBNG workload when co-scheduled with a BE machine learning workload. Moving from 10\% to 20\% BE memory bandwidth allocation (all remaining resources are allocated to the HP workload) has relatively low impact on HP packet drop rate. In contrast, a further increase to 30\% memory bandwidth allocation can incur a 10-100x increase in drop rate (at higher cache allocations). The significant penalties incurred when falling off a QoS cliff have led many prior works to adopt conservative QoS slack thresholds of approximately 10-20\% \cite{Heracles2015, Parties2018, Hipster2017}, thus reducing effective resource utilization.

\begin{figure*}[t]
\centering
\includegraphics[width=0.95\textwidth]{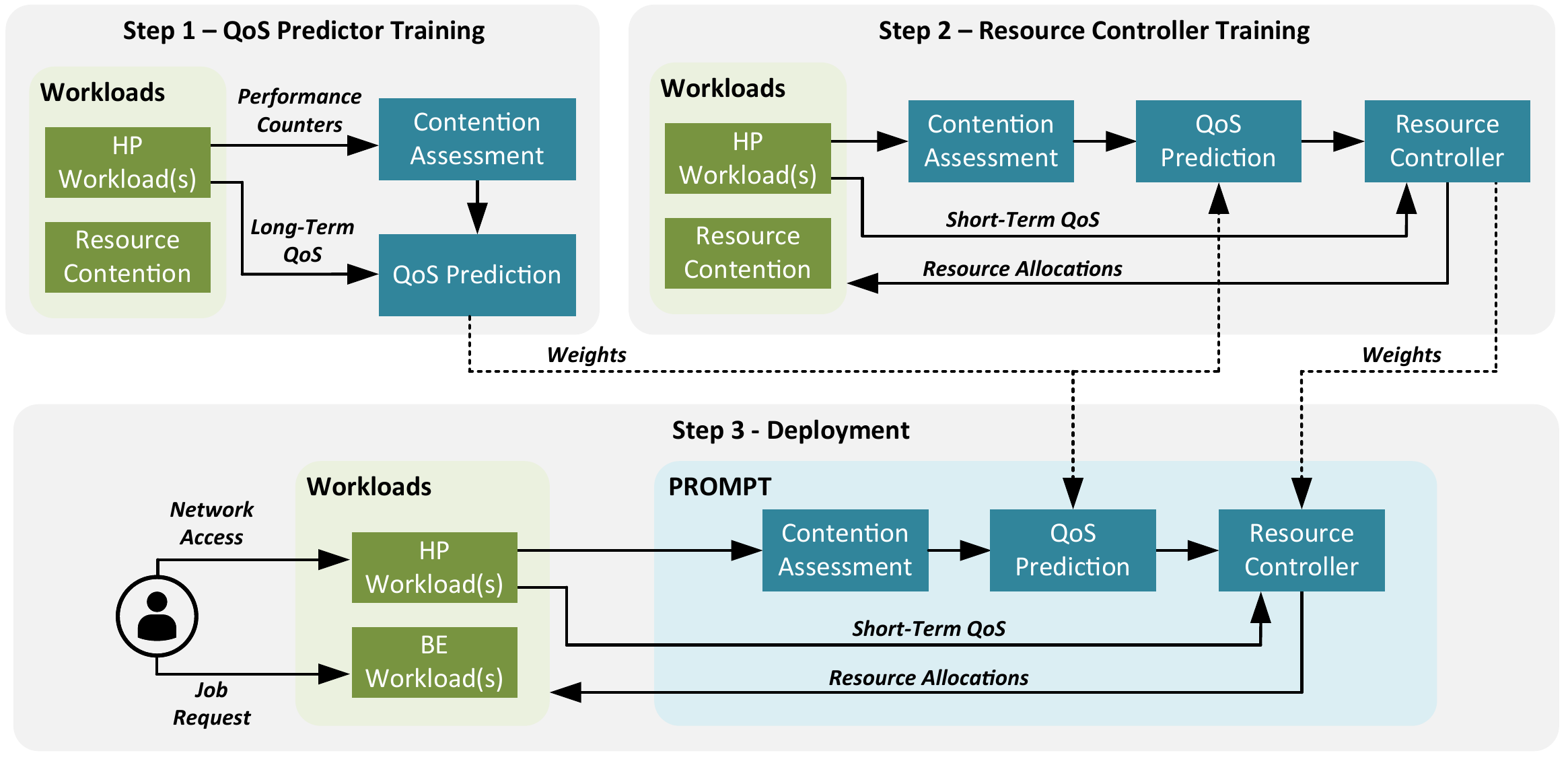}
\caption{PROMPT framework for dynamic resource allocation.}
\label{fig:control_framework}
\end{figure*}

\subsubsection{Transient QoS Violations}
\label{sec:transient_qos_violations}

Prior work has generally focused on HP workloads with latency-based QoS goals that define the maximum latency for 99\% of requests, while allowing arbitrary performance degradation for the remaining 1\% of requests. Given this relatively loose constraint, it is generally practical to estimate worst-case QoS by averaging over brief intervals (e.g., 1-5 seconds) with stable workload demand \cite{Twig2020}. Alternative HP workloads, such as vBNG, require much stricter QoS targets. In fact, industry applications for vBNG commonly target packet drop rates of 0.0001\%. As a result, even brief performance degradations can cause QoS to fall below the target, which we refer to as transient QoS violations. Specifically, as shown in Figure~\ref{fig:coscheduling_behaviors}(b), we observe that the distribution of QoS measurements can exhibit a long upper tail, often over an order of magnitude higher than the average reading. Short measurement intervals can easily miss these critical points with high drop rate, thus providing limited information about worst-case QoS that can result from a particular resource allocation. The presence of these transient behaviors reflects a fundamental limitation in prior studies on QoS prediction, which assumed that QoS can be accurately predicted based solely on workload demand and resource allocation \cite{Adaptive_QoS_Pred_2019, sinan2021}. Consequently, a new methodology is needed to enable general application.

\subsubsection{Generalization to Unseen Behaviors}

The dynamic nature of cloud environments introduces a variety of unknown variables that further complicate resource allocation. In particular, we cannot assume full knowledge of all workloads that may be co-scheduled, so it is impossible to theoretically guarantee whether a particular resource allocation will prevent all QoS violations. Furthermore, there exists significant heterogeneity in platforms (e.g., processors, memory) for various service providers, so any theoretical analysis conducted for a particular system architecture may be invalid for another. Addressing these unknown variables necessitates policies that can, ideally, accommodate a wide range of operating conditions.

Search-based methods typically use QoS measurements as their sole information source to guide resource allocation policy learning. Policies derived from this single, high-level information source cannot be directly related to current workload behaviors, so are expected to become invalid as workload demands change over time. This limitation necessitates repeated online searches, which can be dangerous due to QoS cliffs and may incur significant penalties due to long sampling periods (e.g., 100 seconds) required to accommodate transient behaviors.

Reinforcement-learning-based methods can, instead, use additional information to learn control policies that may remain applicable when workload demands change. Regardless, prior work using reinforcement learning has focused on operating environments with static workloads (i.e., the set of co-scheduled workloads does not change) and none have addressed the challenges created by transient QoS violations. 

\section{PROMPT}
\label{sec:proposed_work}

\begin{table*}[t]
    \centering
    \caption{Selected performance counters (features).}
    \small
    \label{tab:selected_features}
        \begin{tabular}{ccx{7.15cm}}
            \toprule
            &\multicolumn{1}{c}{\textbf{Feature}} & \textbf{Description} \cite{IntelManual} \\
            \midrule
            \multirow{2}{1.65cm}{\textit{Fixed Counters}} & inst\_retired.any & Counts retired instructions \\ 
             & cpu\_clk\_unhalted.thread & Counts cycles when the core is not halted \\ 
          \midrule \multirow{5}{1.65cm}{\textit{General Counters}} & \multirow{2}{*}{frontend\_retired.latency\_ge\_2} & Counts retired instructions following a period of $\geq$ 2 cycles with no uops delivered by the frontend \\ 
             & l2\_rqsts.all\_code\_rd & Counts all code requests at the level-two cache \\ 
             & offcore\_requests.all\_requests & Counts off-core memory transactions \\ 
             & offcore\_requests\_buffer.sq\_full & Counts cases when the off-core requests buffer was full \\
             \bottomrule
		\end{tabular}
\end{table*}

In this work, we propose PROMPT, a deep reinforcement learning framework for dynamic resource allocation guided by proactive QoS predictions. PROMPT is a resource-contention-aware solution for the challenges highlighted in Section \ref{sec:preliminaries}, offering: 1) fine-grained optimization near QoS cliffs, 2) more consistent feedback in the presence of transient QoS fluctuations, and 3) robust generalization when co-scheduling new BE workloads not encountered during policy training. As such, PROMPT enables practical application of dynamic resource allocation in more stringent operating environments.

High-level design for PROMPT is illustrated in Figure~\ref{fig:control_framework}. First, a QoS predictor is trained in an offline environment with synthetic workload demand in order to accurately model transient behaviors over an extended period. Second, the pretrained QoS predictor is used to guide resource controller training by providing estimates of worst-case QoS (e.g., packet drop rate) for each HP workload. Finally, both pretrained models are deployed in the service provider network to provide robust QoS guarantees and improve resource utilization even with previously unseen BE workloads. This framework supports resource allocation between any number of HP and BE workloads, provided that there are sufficient resources to meet QoS goals for all HP workloads.

PROMPT applies several key innovations to address the aforementioned resource allocation challenges. First, resource contention indicators are selected using a hierarchical process that leverages one-time offline profiling of HP workloads (i.e., no BE knowledge required) in order to avoid classes of information sources that generalize poorly. Second, PROMPT accommodates generalization in the specification of state, action, and reward for the resource controller by avoiding problematic assumptions in prior works regarding the number of co-scheduled workloads. Third, QoS predictions made using these low-level resource contention indicators act as the primary feedback source to determine rewards for the reinforcement learning agent. Consequently, consistent control policies can be learned even when co-scheduling workloads with severe transient behaviors. Fourth, PROMPT implements a two-level QoS predictor architecture with multiple regressors, thus allowing each regressor to focus on particular resource contention behaviors and improve the resolution of predictions when operating near the QoS cliff.

\subsection{Assessing Resource Contention}
\label{sec:performance_event_selection}

Modern CPUs implement dedicated counters for architectural events such as cache misses, memory requests, etc. This information can be used to identify deviations from ideal execution behaviors and thereby predict transient QoS behaviors to guide resource allocations. Unfortunately, due to hardware limitations, it is only possible to measure a small subset of these counters without increasing overhead and degrading accuracy, so feature selection is required.

Prior works often selected features manually based on domain knowledge \cite{AIBook}. Naturally, this approach introduces substantial human effort while potentially missing useful counters. PROMPT instead applies a highly-automated hierarchical method that progressively trims the counter list and requires minimal domain knowledge and workload information. Our steps are:
\begin{enumerate}[left=6pt]
  \item We sample each performance counter for one second while running an arbitrary co-scheduling setup. Counters of rare events and counters with low variance are eliminated.
  \item The reduced list of performance counters is sampled for a longer period across a large range of co-scheduling configurations. We then apply BoostARoota \cite{boostaroota} to eliminate all counters that are less informative than randomly permuted readings.
  \item We drop performance counters that exhibit orders-of-magnitude shifts in values when co-scheduling different workloads. Including these counters would likely lead to poor performance when we consider generalization to BE workloads not observed during policy training.
  \item We use domain knowledge to eliminate counters that measure exceptionally specific scenarios or counters that are not easily understood by human experts. 
  \item Finally, we apply step-wise selection methods \cite{Subset_selection} to obtain the desired number of counters.
\end{enumerate}
The final list of performance counters selected via this process is listed in Table \ref{tab:selected_features}. Features listed as ``Fixed Counters'' use separate, dedicated counters, so can be included without overhead.

Note that domain knowledge introduced in step four is not strictly needed for other platforms. In practice, there exist several common classes of potentially problematic counters that can be simply excluded from selection entirely. Specifically, counters relating to memory coherency (e.g., read for ownership requests and snooping) are eliminated since not all workloads involve significant inter-thread dependency. Counters related to specialized instructions sets (e.g., AVX-512) are eliminated as they may not be used by all workloads. Finally, we eliminate counters relating to non-standard architecture components (e.g., loop stream detector), which are only used when specific conditions are met.  

\subsection{Predicting Worst-Case QoS} \label{sec:qos_prediction}

As discussed in Section \ref{sec:transient_qos_violations}, accurate estimates for QoS may require observations over a period of 100 seconds or more when using direct QoS measurements. PROMPT instead leverages QoS prediction in order to shift this extended observation period offline, prior to policy training, and provide reliable QoS estimates even in highly dynamic online operating environments. Specifically, PROMPT trains a QoS predictor to predict the worst HP workload performance $min(P_{meas}(W_{w,t}^{HP}))$ observed over each 100 seconds of offline workload profiling. Online predictions are made based on system state at time $t$, represented as a vector of counter values $\mathbf{c}_{w,t}^{HP}$ for each HP workload $w$, such that $P_{pred}(\mathbf{c}_{w,t}^{HP}) = min(P_{meas}(W_{w,t}^{HP}))$. Training data is gathered across a variety of co-scheduling configurations (i.e., HP workload demands, BE workloads, and resource allocations) in order to improve prediction accuracy under diverse resource contention behaviors.

\begin{figure}[b]
\centering
\includegraphics[width=0.45\textwidth]{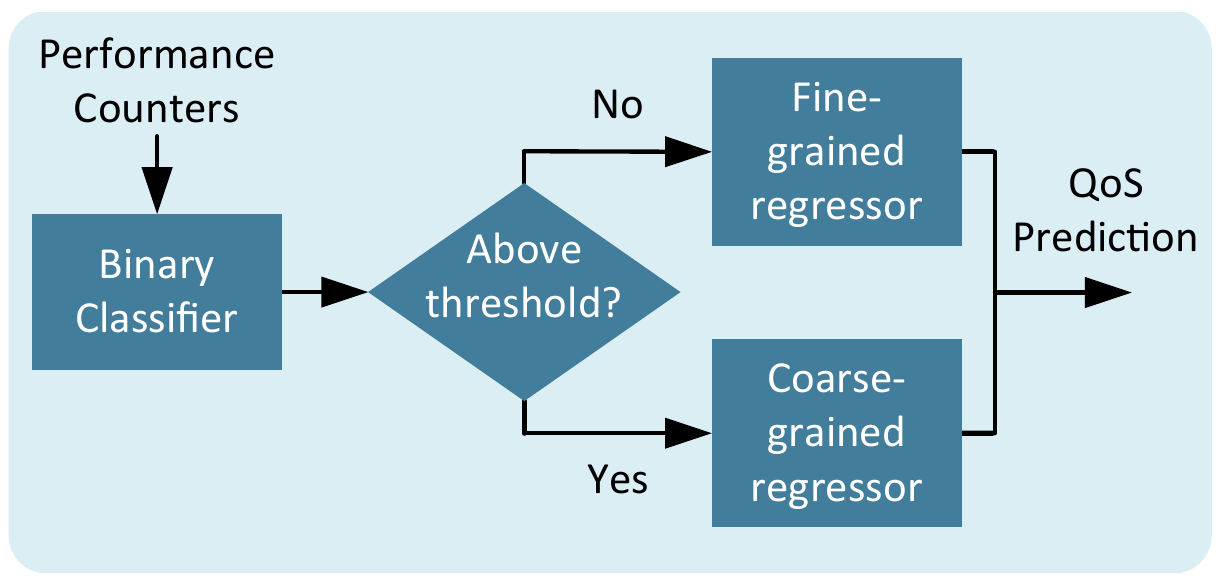}
\caption{QoS predictor architecture.}
\label{fig:predictor}
\end{figure}

\begin{figure}[b]
\centering
\includegraphics[width=0.45\textwidth]{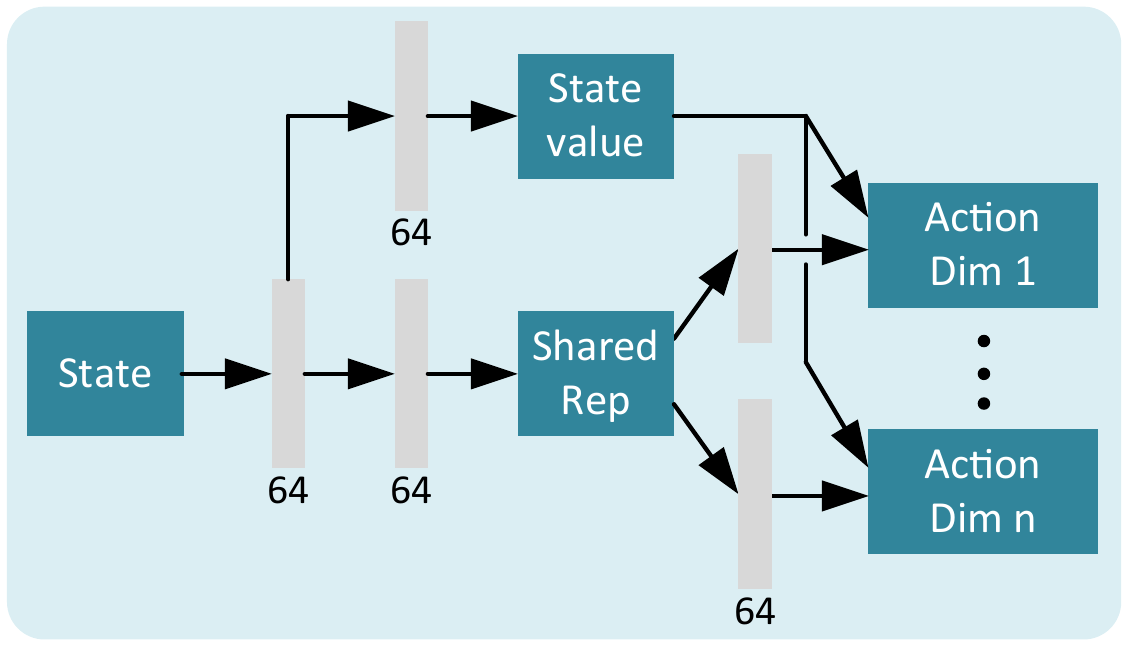}
\caption{Dynamic resource allocation controller architecture.}
\label{fig:bdq}
\end{figure}

Transient QoS behaviors are rare and often extreme events, so directly fitting a regression model on these behaviors can lead to poor QoS predictions. We mitigate these issues using a two-level prediction setup as depicted in Figure~\ref{fig:predictor}. Performance counters are first passed to a binary classifier that predicts whether the worst-case QoS is above or below a specified threshold. This prediction is then used to select between two regressors. The fine-grained regressor is trained on the critical range of QoS values in which the system will ideally operate (e.g., zero up to the threshold value). In contrast, the coarse-grained regressor is trained on the full range of QoS values. By separating these regressors, each can be optimized for particular behaviors that may be encountered during framework execution and improve prediction accuracy. 


\subsection{Learning Optimal Resource Allocation} \label{sec:resource_controller}

We design a resource controller based on deep reinforcement learning that learns a policy $\boldsymbol\pi: \mathbf{S} \rightarrow \mathbf{A}$ mapping system states to resource allocation actions. Our controller specification (i.e., state, action, and reward) accommodates generalized operating environments in which BE workloads may change at any time, thus avoiding frequent and impractical re-training. This generalization introduces various consideration that are not addressed by prior work.

\subsubsection{Model Architecture}
The resource controller in PROMPT is implemented based on the action-branching architecture of Takavoli et al. and their variant of Dueling Double DQN, referred to as Branching Dueling Q-Network (BDQ) \cite{BDQ}. Our controller architecture, depicted in Figure \ref{fig:bdq}, features a shared representation layer, followed by distinct action branches, one for each resource control knob. Splitting these action dimensions allows the number of network outputs to grow linearly, rather than combinatorially, with respect to the action space, thus simplifying control complexity when allocating multiple resources.

\subsubsection{State Specification} 
State specification in PROMPT (Equation \ref{eq:state_spec}) consists of two elements. First, since our reinforcement learning agent inherently requires similar information as the QoS predictor, we re-use HP workload performance counters discussed in Section \ref{sec:qos_prediction}. This re-use is made possible by eliminating performance counters with high distributional shift (as described in Section~\ref{sec:performance_event_selection}). Second, we augment these counters with the \textit{predicted} QoS. The performance counters act as low-level indicators for any deviations in workload execution that may negatively impact QoS. Conversely, the predicted QoS offers a high-level perspective and informs the controller about the margin from the QoS target.
%
\begin{equation}
\label{eq:state_spec}
\mathbf{S} = \{\ \mathbf{c}_{w,t}^{HP} \ , \ P_{pred}(\mathbf{c}_{w,t}^{HP})\ \}
\end{equation}
%

\subsubsection{Action Specification}
Prior work assumed a fixed operating environment in which workloads do not change, so directly selected the resource allocation for all workloads \cite{Twig2020}. In practice, assuming a fixed set of BE workloads is unduly restrictive and causes the controller to become dependent upon the particular BE workloads used during training. PROMPT addresses these issues by indirectly specifying BE resource allocation based on HP resource allocation.

\begin{figure}[tb]
\centering
\includegraphics[width=0.4\textwidth]{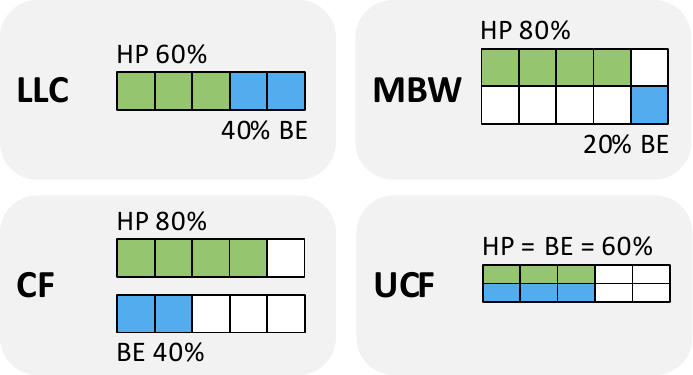}
\caption{Resource allocation action example.}
\label{fig:action_example}
\end{figure}

In this paper, we consider resource allocations defined by four resources $R=\{R_{LLC},R_{MBW},R_{CF},R_{UCF}\}$, including last-level cache (CF), memory bandwidth (MBW), core frequency (CF), and uncore frequency (UCF). Given that the number of BE workloads may change dynamically, we predict allocations only for the HP workload, then determine BE allocations based on HP allocations. We illustrate an example configuration in Figure~\ref{fig:action_example}. Specifically, 
\begin{itemize}[left=6pt]
\item LLC can be strictly divided such that no contention is possible, thus all LLC resources not selected for HP workload(s) are given to BE workload(s). 
We specify a minimum of two LLC units per workload and are free to allocate the remaining units.
\item MBW is not strictly divided but is, instead, allocated on a per-workload basis. As such, contention can be reduced, but not eliminated. Regardless, when co-scheduling, we typically want to prioritize either the HP workload (at high HP demand) or the BE workload (at low HP demand). This effect can be achieved by overlapping the allocation knobs. Values at the upper end of the range prioritize HP MBW while values at the lower end of the range prioritize BE MBW.
\item CF is similar to MBW in that a different setting is possible for all HP cores and BE cores, but CF also introduces power considerations; it may be possible to reduce core frequency for both the HP and BE workload(s) in order to improve power efficiency. We therefore separate these dimensions and select a separate value for HP core frequency (HPCF) and BE core frequency (BECF). 
\item UCF must take the same value for all CPU cores, so we apply the HP workload setting to all HP/BE cores.
\end{itemize}

Allocation options for each resource depend on the target system. In our case, we have 8 possible LLC, 10 MBW, 7 HPCF, 7 BECF, and 5 UCF settings (see \ref{appendix:test_setup}), so resource allocation actions are given as:
\small
\begin{equation}
\mathbf{A}_t \in \{0,...,7\}\times\{0,...,9\}\times\{0,...,6\}\times\{0,...,6\}\times\{0,...,4\}
\end{equation}
\normalsize
We then split these actions across separate prediction heads, such that $\mathbf{A}_t = \{\mathbf{A}_{LLC}, \mathbf{A}_{MBW}, \mathbf{A}_{HPCF}, \mathbf{A}_{BECF}, \mathbf{A}_{UCF}\}$, thus preventing combinatorial explosion.

We have, thus far, discussed a single HP workload for simplicity. Nevertheless, PROMPT is readily extensible to multiple HP workloads by replicating the action space for each HP workload and adding constraints. Specifically, each HP workload could select a number of LLC units, leaving remaining units for the BE workloads. HP MBW would also be separately selected by all HP workloads, while BE MBW would use the lowest setting chosen by HP workloads, essentially restricting the BE to the lowest contention level specified by any HP workloads. The same strategy applies for HPCF and BECF. Finally, UCF would be set to the maximum selected by any HP workload. 

\subsubsection{Reward Specification}
Learning appropriate resource allocation policies necessitates consistent feedback via rewards which, as shown in Section \ref{sec:transient_qos_violations}, is not always possible with QoS measurements alone. PROMPT therefore uses the \textit{predicted} QoS in addition to the \textit{measured} QoS to determine rewards. Specifically, as shown in Equation \ref{eq:r_negative}, negative rewards are given for undesirable actions that cause QoS (either predicted or measured) to degrade beyond the target QoS. Making use of predicted QoS as an additional constraint allows PROMPT to guarantee fewer QoS violations, given sufficient training time, since we can guarantee fewer inappropriate rewards for resource allocations that could lead to QoS violations. In practice, predicted QoS is also a more consistent feedback source since an identical prediction will be made given a fixed workload state. We log transform the QoS ratio since a poor action selection can lead to predicted/measured QoS being several orders-of-magnitude greater than the target QoS, which could increase difficulty in policy learning across various co-scheduling scenarios, similar to a multi-game environment \cite{dqn}. Finally, we clip this value at $\beta$, acting as an upper threshold at which it is expected to be beneficial to differentiate violation severity. We empirically determined $\beta=3$ to be appropriate for workloads in evaluation, given that violations more severe than this threshold (i.e., qos $\ge$ 1000x the target) likely all require maximal throttling to quickly re-establish an appropriate QoS, so would only introduce noise in Q-value estimates.
\begin{equation}
\label{eq:r_negative}
r_{w}^{-} = -\min\big[log\left(\dfrac{\max(\ P_{pred}(\mathbf{c}_{w,t}^{HP})\ ,P_{meas}(W_{w,t}^{HP})\ )}{Q_w^{HP}}\right),\,\beta\,\big]
\end{equation}
Positive rewards are given when the predicted QoS is less than the target QoS. In this case, we consider our secondary goals of optimizing BE performance while reducing power consumption. These goals are specified in Equation \ref{eq:r_positive}. BE performance $P_{meas}(W_{w,t}^{BE})$ can be increased by increasing LLC, MBW, BE core frequency, or uncore frequency. Overall power consumption, denoted as $C_{meas}(W_{t})$, can be reduced by decreasing HP core frequency, BE core frequency, or uncore frequency. We balance these goals with the parameter $\alpha$, which we set to be 0.8, thus favoring BE performance. Both performance and power consumption are normalized [0,1].
\begin{equation}
\label{eq:r_positive}
r_{w}^{+} = \alpha * P_{meas}(W_{w,t}^{BE}) + (1-\alpha) * C_{meas}(W_{t})
\end{equation}
%

\section{Evaluation}\label{sec:results}
\label{sec:methodology}

\subsection{Methodology}
\subsubsection{Prior Works for Comparison}
Prior works, as originally proposed, made various problematic assumptions that preclude meaningful comparison with PROMPT. The state-of-the-art search-based method, CLITE \cite{CLITE2020}, assumed relatively static workload demands and did not define conditions under which re-sampling should be initiated. Similarly, the state-of-the-art reinforcement-learning-based methods assumed either a static set of workloads \cite{RLDRM2020} or did not support BE workloads \cite{Twig2020}. In contrast, only PROMPT accommodates more generalized operating environments with highly-dynamic resource demands and varying mixes of previously unseen BE workloads. For comparison, we therefore implement enhanced versions of the closest prior works, CLITE and Twig, that support: 1) both HP and BE workloads with varying demand, 2) allocation for all resources supported by PROMPT, 3) packet-based QoS, and 4) power goals. We refer to these versions as CLITE+ and Twig+. These modifications and model parameters are detailed in \ref{appendix:test_setup}.

\subsubsection{Platform}
Experiments were conducted on an Intel\textsuperscript{\textregistered} Xeon\textsuperscript{\textregistered} Platinum 8280. All cores used in testing were isolated using the \textit{isolcpus} kernel option to avoid interference from any other tasks. Workloads were pinned to distinct subsets of these isolated cores. As with most prior work, we used these distinct subsets (no overlapping) to prevent any contention in the private cache for each core. Turbo boost technology was disabled.

\subsubsection{Workloads}
Evaluation focuses on the HP vBNG workload given its prominent role in networking applications \cite{BNG_whitepaper}. We also demonstrate PROMPT's general applicability to other HP workloads (e.g., 5G UPF \cite{5G_UPF_whitepaper, 5G_UPF_announcement, aether_summary, 5G_UPF_code}). Both the vBNG and the 5G UPF workload implement multi-stage packet processing pipelines for traffic to and from service provider networks. Pipeline stages include packet integrity verification, access restriction checks, packet classification, usage metering, and routing determination \cite{BNG_whitepaper, 5g_upf_pipeline}. Load for HP networking workloads is generated on an external server running TRex traffic generator \cite{trex} and follows a diurnal load cycle measured in a real-world network gateway. BE workloads were selected from industry-standard benchmark suites, including SPEC CPU2006 \cite{SPEC_CPU2006}, SPEC CPU2017 \cite{SPEC_CPU2017}, PARSEC \cite{PARSEC}, and several machine learning workloads \cite{TF_models}. Selection criteria for BE workloads was based on sensitivity to resource allocation while maintaining a diverse set of workloads from many domains, representing data compression (bzip2), code compilation (gcc), fluid dynamics (lbm), computational electromagnetics (fotonik3d), and more.

\subsubsection{Train/Test Setup}
QoS predictor training data comprises roughly 500 examples for each of nine co-scheduled workload mixes (HP + BE), with the worst-case QoS measured over a 100-second period. Predictor accuracy is tested using k-fold cross validation in which each fold comprises the 500 examples from a particular co-scheduled workload mix. Similarly, for control evaluation, we use a QoS predictor model that is trained without any data from the currently co-scheduled BE workload(s). The reinforcement learning model in both Twig+ and PROMPT is trained with STREAM \cite{STREAM1995} and SPEC JBB2005\footnote{STREAM serves as a ``worst-case'' co-scheduled workload due to its high-intensity memory bandwidth usage. JBB2005 is a representative workload specified by service providers.} as the \textit{only} BE workloads; neither model has \textit{any} prior knowledge of the 22 BE workloads used in evaluation.

\begin{table}[t]
    \centering
    \caption{QoS Prediction Comparison. ``Select'' refers to features from Section \ref{sec:performance_event_selection}. ``Loose'' refers to a looser classifier threshold. }
    \small
    \label{tab:prediction_accuracy}
        \begin{tabular}{lccc}
            \toprule
            \multirow{2}{*}{\textbf{Configuration}} & \textbf{Classifier} & \multicolumn{2}{c}{\textbf{Regressor (MAE)}} \\
            \cmidrule(lr){2-2} \cmidrule(lr){3-4}
            & F$_{1}$ score & Critical range & Full range \\
            \midrule
            (1) 1-level, IPC & N/A & 75.15 & 302.36 \\
            (2) 2-level, IPC & 0.853 & 42.97 & 302.43 \\
            (3) 1-level, Select & N/A & 26.52 & 162.95 \\
            (4) 2-level, Select & 0.908 & 17.01 & 161.19 \\
            (5) 2-level, Select, Loose & 0.866 & 20.07 & 162.42 \\
            \bottomrule
		\end{tabular}
\end{table}

\subsection{Results}
\subsubsection{QoS Prediction Accuracy}
We evaluated our QoS predictor against several baselines, as shown in Table~\ref{tab:prediction_accuracy}. These baselines use either instructions-per-cycle (IPC) only, one-level QoS prediction (i.e., a single regressor), or both. Config 2 and 4 use a binary classifier threshold of 250 dropped packets per second (dpps) while config 5 uses a threshold of 2500 dpps. The simplest setup (config 1) achieves a mean-average-error (MAE) of 75.15 dpps in the critical range (0 to QoS Target), which provides limited improvement over individual QoS measurements given a scenario similar to that in Figure~\ref{fig:coscheduling_behaviors}(b). Adding two-level prediction (config 2) significantly reduces regression error in the critical range, even when using IPC only; resource contention behaviors tend to be similar in this critical range, thus simplifying the task of the fine-grained regressor. Alternatively, adding the selected features (from Section \ref{sec:performance_event_selection}) to a one-level predictor (config 3) further reduces regression error to 26.52 dpps, indicating that IPC alone is not sufficient to accurately predict QoS. Finally, adding the selected features to the two-level setup (config 4) provides the best result in all metrics. The benefit of two-level prediction can also be quantified by the number of large mispredictions, where regression error exceeds 50 dpps. Moving from a one-level predictor (config 3) to a two-level predictor (config 4) eliminates 46.5\% of large mispredictions (from 254 to 136). Finally, we examine the impact of the classifier threshold. Loosening this threshold (config 5) degrades classifier F1 as it is more difficult to differentiate severe contention behaviors. Critical range regression error also degrades as the total number of large mispredictions increases by 34.5\%. Consequently, we use config 4 (threshold approximately equal to the QoS target) for all further experiments.

\begin{figure*}[b]
\centering
\includegraphics[width=0.99\textwidth]{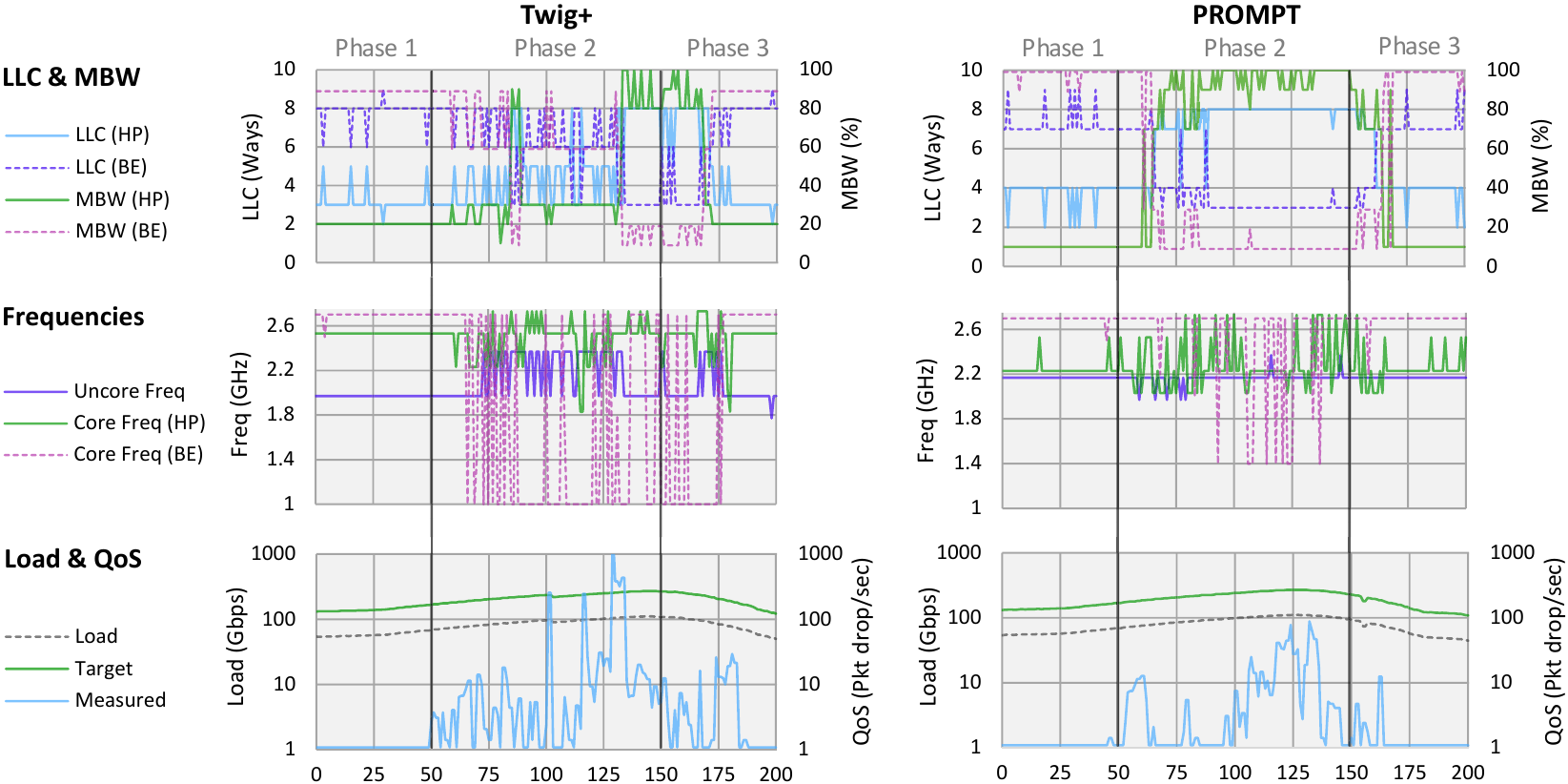}
\caption{Behavior comparison when co-scheduling vBNG and cactuBSSN (SPEC) workloads. The top two rows illustrate the policy actions while the bottom row illustrates HP workload demand and QoS (target and measured).}
\label{fig:control_behavior_vbng}
\end{figure*}

\subsubsection{Practical Control Capabilities} 
We first tested whether CLITE+, which represents the state-of-the-art in search-based methods, is suitable for highly-dynamic operating environments with strict QoS guarantees. For this test, we co-scheduled the HP vBNG workload alongside a machine learning training workload which is known to cause modest resource contention. HP load follows a detailed traffic trace gathered from a real-world operating environment. The total test duration (four hours) was chosen such that CLITE+ could always successfully finish sampling (i.e., no significant changes in workload demand during sampling). Even with this simplified operating environment, we observed dramatic differences in QoS violations between CLITE+ and PROMPT. As shown in Table \ref{tab:clite_comparison}, the online search performed by CLITE+ caused 14 occurrences (roughly 42 seconds) during which over one million packets were dropped per second, roughly 10\% of the \textit{total} injected traffic. These periods could cause substantial interruptions in real-time services so are unacceptable in many operating environments. In contrast, the worst-case QoS violations incurred by PROMPT corresponded to roughly 0.1\% of the total traffic per second. These violations were caused by temporary reductions in HP resource allocation during a period of high HP demand. These problematic actions were, however, extreme outliers; practically all nearby resource allocation actions were appropriate and nearly identical. These violations can therefore be eliminated by a time-window average over recent actions, as demonstrated in later subsections. Conversely, CLITE+ cannot apply windowed averaging to address systematic violations during sampling. Furthermore, these issues would be exacerbated in more dynamic operating environments that require CLITE+ to re-sample more often. Given these limitations, we did not include CLITE+ in further tests.

\begin{table}[t]
    \centering
    \caption{QoS comparison. PROMPT has fewer intervals (count) with high drop rate and lower average drop rate.}
    \small
    \label{tab:clite_comparison}
        \begin{tabular}{lx{1.3cm}x{1cm}x{1.1cm}x{1.1cm}x{1cm}}
            \toprule
            \multirow{2}{0.6cm}{\textbf{Dpps range}}  &  & \multirow{2}{*}{0-1K} & \multirow{2}{*}{1K-10K} & \multirow{2}{*}{10K-1M} & \multirow{2}{*}{$>$1M} \\
            \\
            \midrule
            \multirow{2}{0.6cm}{\textbf{Count}} & \multirow{1}{*}{CLITE+} & \multirow{1}{*}{4806} & \multirow{1}{*}{50} & \multirow{1}{*}{112} & \multirow{1}{*}{14} \\
            & \multirow{1}{*}{PROMPT} & \multirow{1}{*}{4935} & \multirow{1}{*}{40} & \multirow{1}{*}{7} & \multirow{1}{*}{0} \\
            \midrule
            \multirow{2}{0.6cm}{\textbf{Avg dpps}} & CLITE+ & 34 & 4,241 & 228,136 & 1.83M\\ 
            & PROMPT & 17 & 2,464 & 18,330 & 0 \\ 
            \bottomrule
		\end{tabular}
\end{table}

\begin{table*}[t]
\centering
\caption{Co-scheduling results across 22 BE workloads. We report arithmetic mean for QoS violation \% and geometric mean for other metrics. Workload abbreviations: bwa = bwaves; bzi = bzip2; caa = cactusADM; cab = cactuBSSN; fac = facesim; flu = fluidanimate; gcc = gcc; gem = gemsFDTD; ima = image classifier training (ResNet); lbm = lbm; les = leslie3d; mcf = mcf; mil = milc; omn = omnetpp; par = parest; rom = roms; sig = signal classifier training (boosted trees); sop = soplex; sph = sphinx3; wrf = wrf; zeu = zeusmp.}
\label{tab:summary}
\includegraphics[width=1.0\textwidth]{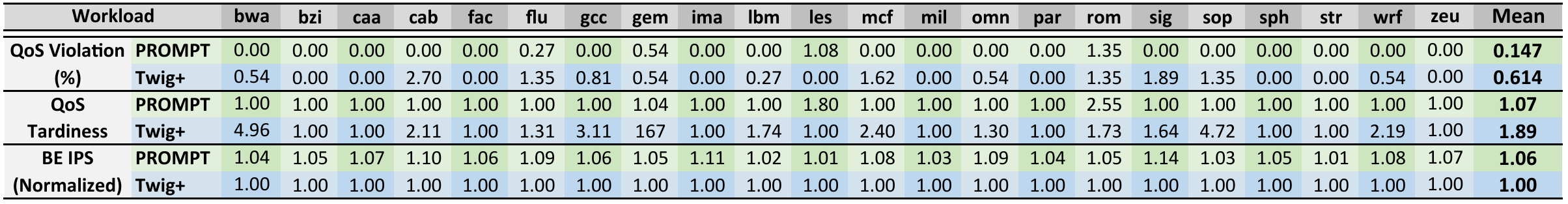}
\end{table*}

\begin{figure*}[t]
\centering
\includegraphics[width=0.99\textwidth]{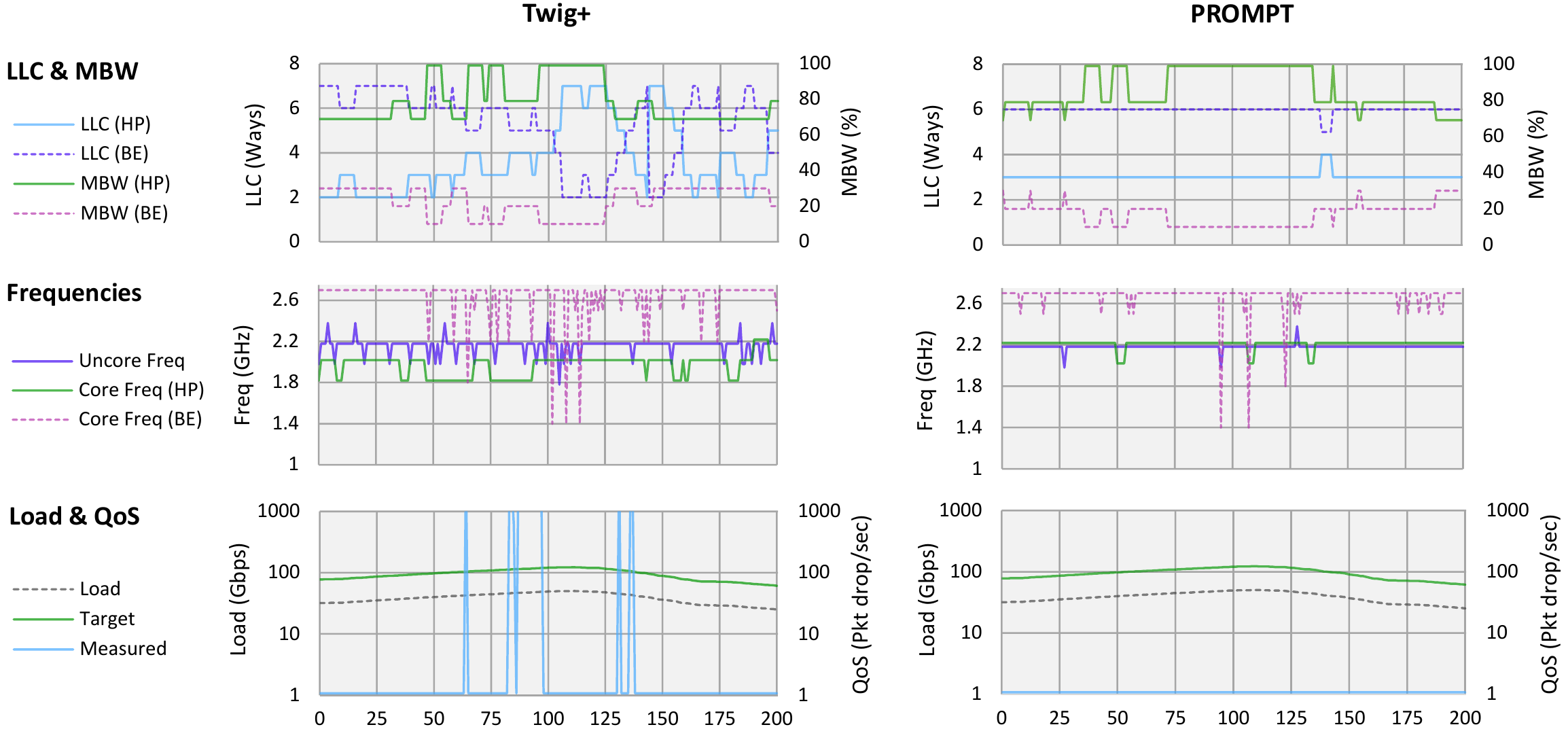}
\caption{Behavior comparison when co-scheduling 5G UPF and lbm (SPEC) workloads. The top two rows illustrate the policy actions while the bottom row illustrates HP workload demand and QoS (target and measured).}
\label{fig:control_behavior_upf}
\end{figure*}

\subsubsection{Resource Allocation Behaviors}
Policy learning guided by QoS predictions (PROMPT), rather than QoS measurements only (Twig+), improves the consistency of resource allocation decisions in real-world evaluation. As shown in Figure~\ref{fig:control_behavior_vbng}, PROMPT correctly identifies periods of low HP load (phase 1), during which few resources are allocated to the HP. As the HP load rises (phase 2), PROMPT briefly fluctuates between several LLC/MBW options and then settles on a relatively stable configuration. PROMPT continues to make minor BE core frequency adjustments, when necessary, to mitigate transient QoS fluctuations. Finally, when HP load decreases (phase 3), resources are again taken from the HP and given to the BE. In contrast, Twig+ exhibits relatively inconsistent behavior during phase 2 as it cycles between overly aggressive actions (causing a violation) and more conservative actions (recovering from a violation). Most violations incurred by Twig+ followed a series of complex actions in which Twig+ simultaneously adjusted \textit{every} resource. These behaviors are difficult to mitigate, even with a time-window average over policy decisions, since Twig+ has less knowledge about transient behaviors.

\subsubsection{Resource Allocation Efficacy}
We evaluated the policies learned by PROMPT and Twig+ when co-scheduling the vBNG workload with each of 22 BE workloads not seen during policy training. Results are shown in Table \ref{tab:summary}. First, we compare the number of QoS violations incurred by each method. Only 0.147\% of actions made by PROMPT result in QoS violations, compared with 0.614\% for Twig+ (a 4.2x increase). These percentages may seem small, but are crucial in industry applications with strict service level agreements. In practice, the higher proportion of QoS violations incurred by Twig+ could lead to substantial penalties, potentially millions of dollars, and higher operational costs for customer support and maintenance. Next, we compare the QoS tardiness,\footnote{The ratio of measured to target when measured exceeds target.} which quantifies the severity of QoS violations. On average, the QoS violations incurred by Twig+ were 1.89x the target whereas the QoS violations incurred by PROMPT were just 1.07x the target. In other words, the QoS violations incurred by Twig+ were 12.7x more severe. Although improving HP QoS usually requires more restrictive BE resource allocation, thus reducing BE performance, PROMPT actually improved BE performance. Specifically, PROMPT improved average BE performance (measured as instructions-per-second) by an average of 6\% (and up to 14\%) compared with Twig+, since Twig+ incurs significant penalties when recovering from QoS violations. Finally, PROMPT improved average power efficiency by 5\% compared with Twig+. PROMPT therefore offers better results in all metrics by incurring fewer and less severe QoS violations while still improving BE performance.

\subsubsection{Application to Other HP Workloads}
Transient QoS behaviors (Section \ref{sec:preliminaries}) are not limited to the vBNG workload. In fact, we observed similar behaviors across multiple networking workloads, including the 5G UPF --- a critical workload that connects mobile users to a service provider network. When co-scheduling this workload, we observed scenarios with severe transient behaviors in which QoS measurements fluctuated between 0 dpps and 10K+ dpps, even with constant resource allocation and HP load. We further verified that these transient behaviors can be predicted and, therefore, mitigated by PROMPT. As shown in Figure \ref{fig:control_behavior_upf}, we again see that PROMPT tends to learn more consistent resource allocation policies with stable QoS while the overly aggressive policies learned by Twig+ result in periods with significant QoS degradation. Overall, across all 22 BE workload combinations, Twig caused QoS violations on 0.70\% of actions and these violations were 4.64x the target, on average. In contrast, PROMPT caused QoS violations on just 0.024\% of actions and these violations were 1.13x the target. In other words, PROMPT incurred 29x fewer QoS violations and reduced the severity of these violations by 28x. Finally, PROMPT improved BE performance (0.3\%).

\subsubsection{Controller Overhead}
Twig+ and PROMPT use an identical reinforcement learning model, thus execution time is identical except for overhead due to QoS prediction. Without QoS prediction, each control interval requires approximately 12ms when running on one core. QoS prediction incurs an additional 1ms overhead. This additional overhead is easily outweighed by the gains in BE performance. Note that both Twig+ and PROMPT are executed every three seconds during evaluation, thus total execution overhead is around 0.5\% on a single core.

\section{Conclusion}

Co-scheduling of high-priority and best-effort workloads, enabled by dynamic resource allocation, can greatly improve server utilization and reduce total cost of ownership. Practical application, however, requires strict mitigation of diverse resource contention behaviors, several of which are not adequately handled by prior work. PROMPT addresses these issues with a generalized framework based on proactive QoS prediction, thereby enabling more precise resource optimization, more consistent handling of transient performance fluctuations, and more robust generalization when co-scheduling new BE workloads not encountered during policy training. Evaluation shows that the proposed framework incurs 4.2x fewer QoS violations, reduces the severity of QoS violations by 12.7x, improves BE performance, and improves power efficiency compared with prior work.


\section*{Acknowledgments}
This work was supported by Intel Corporation's Academic Research funding. 

\appendix
\section{Detailed Test Setup} \label{appendix:test_setup}

\subsection{QoS Predictors}
All QoS predictors (classifiers and regressors) were implemented as boosted decision trees, with varying depth and number of trees. The classifier used 20 estimators with a depth of 2. The coarse-grained regressor used 20 estimators with a depth of 3 while the fine-grained regressor used 30 estimators with a depth of 3. We used a learning rate of 0.2 for all models, which helped to reduce the number of estimators and thereby lower inference overhead.

\subsection{Reinforcement Learning Models (Twig+ and PROMPT)}
The original Twig framework was designed for a very specific operating environment, so significant modifications were required for meaningful comparison. Briefly, Twig supported only HP workloads, latency-based QoS, and only allocated cores and uncore frequency. Our generalized framework (excluding QoS prediction) described in Section \ref{sec:proposed_work} is therefore used for both Twig+ and PROMPT. We also log transform and normalize the input state based on the distribution for QoS prediction data.

The deep Q-network in both Twig+ and PROMPT is based on an action-branching architecture (described in Section \ref{sec:resource_controller}) with additional branches to accommodate separate state and action dimensions for each HP workload (similar to Twig \cite{Twig2020}). All network layers have 64 nodes, which we found to achieve a good balance between overhead and learning capacity. Models are trained for a maximum of 25,000 steps using a three-second interval. Training uses dueling networks with target network updates every 100 steps, prioritized experience replay ($\alpha$=0.6, $\beta$=0.4, $\epsilon$=1e-6, batch size = 256), and an Adam optimizer (initial learning rate = 0.002). Weights are saved every 500 steps. Weights used for evaluation were selected to minimize the running-average number of QoS violations. Epsilon-greedy exploration was used during training (decaying from 100\% to 2\%) and then disabled for evaluation. Both Twig+ and PROMPT are implemented with Tensorflow. All testing assumes that Twig+ and PROMPT can be trained offline. If the target operating environment cannot be adapted for offline training, it may be necessary to consider further improvements (e.g., transfer learning from an operating environment that can be adapted for offline training).  

\subsection{Bayesian Optimization Configuration (CLITE)}
CLITE \cite{CLITE2020} was originally designed as a one-off search that simply returned the most optimal resource allocation found during exploration. We therefore made several changes to accommodate continuous operation (in addition to those described in Section \ref{sec:methodology}). First, CLITE+ continuously measures the QoS for each HP workload and re-samples whenever more than 3 QoS violations are recorded in the past 10 measurement intervals. This mainly affects operation when HP workload demand is increasing. Second, CLITE+ continuously measures the workload demand for each HP workload and re-samples whenever the observed workload demand has dropped by more than 5\% since the end of the previous sampling period. These constraints cause CLITE+ to perform a full search roughly ten times per day if we only consider changes in HP workload demand (with a diurnal load profile).

\subsection{Detailed Action Specification}
As mentioned in Section \ref{sec:resource_controller}, action specification is largely guided by platform resources. Here, we explicitly detail the action steps used in our testing. LLC is allocated in steps of 1 way; with a maximum of 7 ways, we can give the HP $x\in\{0,...,7\}$ ways and all remaining ways are given to the BE, resulting in 8 possible actions. MBW actions are specified by the maximum allowed allocation. This allocation is specified with 10\% granularity from 10\% to 100\%, resulting in 10 possible actions. Granularity for core frequency and uncore frequency steps is highly platform dependent and only certain frequency steps may be supported. In order to reduce action space complexity, we select a relatively coarse-grained subset of steps from those supported by our platform. This subset includes $CF\in\{1.0, 1.4, 1.8, 2.0, 2.2, 2.5, 2.7\}$ and $UCF\in\{1.6, 1.8, 2.0, 2.2, 2.4\}$, which correspond to 7 possible HPCF/BECF actions and 5 possible UCF actions.


\bibliographystyle{elsarticle-num} 
\bibliography{refs}





\end{document}